\relax
\documentclass[pmlr]{jmlr}

\usepackage{longtable}
\usepackage{booktabs}
\usepackage[load-configurations=version-1]{siunitx} 

\theorembodyfont{\upshape}
\theoremheaderfont{\scshape}
\theorempostheader{:}
\theoremsep{\newline}

\jmlrvolume{1}
\jmlryear{2021}
\jmlrworkshop{AAAI Spring Symposium 2021 (SP-ACA)}

\title[Deep-CR MTLR]{Deep-CR MTLR: a Multi-Modal Approach for Cancer \titlebreak Survival Prediction with Competing Risks}
\author{
    \Name{Sejin Kim} \textsuperscript{1, 2, *} \Email{sejin.kim@uhnresearch.ca}\\
    \Name{Michal Kazmierski}\textsuperscript{1, 2, *} \Email{michal.kazmierski@uhnresearch.ca}\\
    \Name{Benjamin Haibe-Kains}\textsuperscript{1, 2} \Email{benjamin.haibe-kains@uhnresearch.ca}\\
    \addr \textsuperscript{\rm 1} Department of Medical Biophysics, University of Toronto, Toronto, Ontario, Canada.\\
    \addr \textsuperscript{\rm 2} Princess Margaret Cancer Centre, University Health Network, Toronto, Ontario, Canada.\\
    \textsuperscript{\rm *} These authors contributed equally.
}

\begin{document}
\maketitle

\begin{abstract}
Accurate survival prediction is crucial for development of precision cancer medicine, creating the need for new sources of prognostic information. Recently, there has been significant interest in exploiting routinely collected clinical and medical imaging data to discover new prognostic markers in multiple cancer types. However, most of the previous studies focus on individual data modalities alone and do not make use of recent advances in machine learning for survival prediction. We present \textit{Deep-CR MTLR} --- a novel machine learning approach for accurate cancer survival prediction from multi-modal clinical and imaging data in the presence of competing risks based on neural networks and an extension of the multi-task logistic regression framework. We demonstrate improved prognostic performance of the multi-modal approach over single modality predictors in a cohort of 2552 head and neck cancer patients, particularly for cancer specific survival, where our approach achieves 2-year AUROC of 0.774 and \(C\)-index of 0.788.

\end{abstract}

\section{Introduction}
Over 18 million patients were diagnosed with cancer worldwide in 2018. Head and neck cancer (HNC) is the 7th most common type of cancer, with 890,000 diagnoses and over 450,000 deaths \cite{bray_global_2018}. The outcomes in HNC have been improving \cite{longo_head_2020}, and being able to accurately predict the patient's prognosis is crucial for selecting optimal treatment to balance the odds of cure against the risk of relapse and treatment side effects \cite{lambin_predicting_2013}. Currently, clinicians rely on standard prognostic factors such as tumour stage, which provide only coarse stratification and are insufficient for the purpose of precision medicine \cite{lambin_predicting_2013}.

Currently, clinicians use non-invasive medical imaging, such as computed tomography (CT) to assess high-level, qualitative features like the size and spread of the primary tumor and lymph node metastases to evaluate patients according to the staging criteria \cite{bray_global_2018}. However, these images contain quantitative information that is not captured by the semantic features used in staging. Multiple recent studies have been conducted trying to discover prognostic imaging biomarkers using both hand-engineered features in combination with traditional machine learning algorithms \cite{aerts_decoding_2014} and deep neural networks \cite{hosny_deep_2018} under the umbrella term of \textit{radiomics}. In HNC, both hand engineered CT features and deep convolutional neural networks (convnets) on CT images were previously used to predict overall survival (i.e. death from any cause) \cite{vallieres_radiomics_2017}. These past approaches often focus on imaging data alone and do not combine it with other available data modalities, such as electronic medical records (EMR), or integrate them only in an ad-hoc fashion. They also either do not use dedicated survival prediction methods, replacing them with proxies like binary classification, or use models relying on overly restrictive assumptions (e.g. linearity and proportional hazards). Crucially, none of the prior studies considers the problem of \textit{competing risks} --- the possibility of multiple mutually-exclusive types of event. This is especially important in cancer prognosis, where accurate estimates of failure risk from primary disease and other causes (e.g. due to comorbidities) are necessary to balance treatment effectiveness and toxicity. Additionally, it is intuitively clear that incorporating cancer imaging should help the most in predicting cancer specific death and less so for unrelated causes. Conflating different causes of death in one overall survival outcome, as has been done previously, might make it harder to learn discriminative image features and obscure the true added value of imaging.

We propose \textit{Deep-CR MTLR} which uses deep neural networks to learn joint prognostic representations between CT images and clinical information, and a competing risks extension of multi-task logistic regression to accurately predict cancer-specific and other cause survival. We test the model on a large cohort of HNC patients and show improved prognostic performance for 2-year and lifetime cancer death risk prediction over the individual data modalities alone.

\section{Methods}
\subsection{Survival Analysis With Competing Risks}
\textit{Survival analysis} is concerned with estimating the time to event (e.g. death from cancer), often from incomplete observations (a problem known as \textit{censoring})  \cite{kalbfleisch_statistical_2002}. In this study, we consider the problem of \textit{competing risks} --- survival prediction when more than one event is possible, and each patient only experiences one of the events. This is a common scenario in cancer prognosis, where a patient can die either of primary cancer or other complications (e.g. cardiovascular disease). Formally, given a training dataset \(D = \{T^{(j)}, e^{(j)}, \mathbf{x}^{(j)}\}_{j=1}^{N}\), where \(T^{(j)}\) is the time to event or censoring for patient \(j\), \(e^{(j)}=0,1,\dots,E\) is the event indicator (with 0 denoting censoring) and \(\mathbf{x}^{(j)}\) are the patient features, we want to learn to predict the probability that an event occurs before some time \(t\), \(P(T^{(j)} \le t, e^{(j)}=e \mid \mathbf{x}^{(j)})\), known as the \textit{cumulative incidence function} (CIF) for event \(e\). Unless the competing risk are independent, which is generally not the case, using a standard survival prediction model on each event separately will give biased probability estimates \cite{kalbfleisch_statistical_2002} , which warrants the use of dedicated competing risks methods.

\subsection{Multi-Task Logistic Regression For Competing Risks Analysis}
Multi-task logistic regression (MTLR) is a machine learning method for individual survival prediction, first proposed by \cite{yu_learning_2011} in single-event setting (i.e. for \(E = 1\)). It casts the problem as a multi-task binary classification on discretized time axis and uses a sequence of dependent logistic regressors to ensure consistency of predictions (i.e. that no patient experiencing an event 'comes back to life' at a later timepoint). Specifically, the time axis is divided into \(K\) discrete intervals, \(c_k = (t_{k-1}, t_k]\), with \(t_0=0\) and \(t_{K}=\infty\). Defining \(\mathbf{y}\) as a sequence of binary random variables \((Y_1, ..., Y_K)\), \(Y_k = \mathbf{1}\{t_{k-1} < T \le t_k\}\), where \(\mathbf{1}\{\}\) is the indicator function (equal to 1 when the event occurs in the interval \(c_k\) and 0 otherwise), we can write the predicted probability mass function (PMF) as 
\begin{align}
\begin{split}
P_{\boldsymbol{\Theta}}(t_{k-1} < T \le t_k \mid \mathbf{x}) &= P_{\boldsymbol{\Theta}}(\mathbf{y} = (y_1, \dots, y_{K-1}) \mid \mathbf{x})\\
&= \frac{1}{Z(\boldsymbol{\Theta}, \mathbf{x})}\exp(\sum_{k=1}^{K-1}(\boldsymbol{\theta}_k^T \mathbf{x} + b_k)y_k),
\end{split}
\end{align}
where \(\boldsymbol{\Theta} = \{\boldsymbol{\theta}_k, b_k\}_{k=1}^{K-1}\) are the model parameters in each time interval and \(Z(\boldsymbol{\Theta}, \mathbf{x}) = \sum_{i=1}^{K}\exp(\sum_{k=i}^{K-1}\boldsymbol{\theta}_k^T
\mathbf{x} + b_k)\) is the normalizing constant. 

Right censoring is addressed by marginalizing over the unobserved intervals past the censoring time: \begin{small}
\begin{align}
P_{\boldsymbol{\Theta}}(T > t_j \mid \mathbf{x})
&= \frac{1}{Z(\boldsymbol{\Theta}, \mathbf{x})}\sum_{i=j}^{K-1}\exp(\sum_{k=i}^{K-1}(
\boldsymbol{\theta}_k^T \mathbf{x} + b_k)y_k),
\label{eq:mtlr_censoring}
\end{align}
\end{small}
with \(t_j\) denoting the time of censoring. The MTLR log-likelihood for a dataset \(D\) with \(N\) patients, \(N_c\) of whom are censored, is then given by
\begin{small}
\begin{align}
L(\boldsymbol{\Theta}, D) &=\sum_{j=1}^{N-N_c}\sum_{k=1}^{K-1}(\boldsymbol{\theta}_k^T \mathbf{x}^{(j)} + b_k)y_k^{(j)} &\text{(uncensored)}\nonumber\\
    &+ \sum_{j=N-N_c+1}^{N}\log(\sum_{i=1}^{K-1}\mathbf{1}\{t_i \ge T_c^{(j)}\}\exp(\sum_{k=i}^{K-1}(\boldsymbol{\theta}_k^T
    \mathbf{x}^{(j)} + b_k)y_k^{(j)})) &\text{(censored)}\\
    &- \sum_{j=1}^{N}\log(Z(\boldsymbol{\Theta}, \mathbf{x}^{(j)})) &\text{(normalizing constant)}\nonumber,
\label{eq:mtlr_likelihood}
\end{align}
\end{small}
where \(T_c^{(j}\) is the time of censoring of \(j\)th instance.

To extend the MTLR framework to competing risks, we introduce a separate set of parameters \(\boldsymbol{\Theta}_e = \{\boldsymbol{\theta}_{e,k}, b_{e,k}\}_{k=1}^{K-1},\ 1 \le e \le N_e\) for each of the \(N_e\) competing events. We define \(Y_{e,k} = \mathbf{1}\{t_{k-1} < T \le t_k \land E = e\}\)
to be the indicator for event \(e\) at time \(t_k\) and compute the predicted joint probability of event and time as
\begin{align}
\begin{split}
\hat P_{\boldsymbol{\Theta}}(t, e, \mathbf{x}) &\equiv
\hat P_{\boldsymbol{\Theta}}(t_{k-1} < T \le t_k, E = e \mid \mathbf{x})\\
&= \hat P_{\boldsymbol{\Theta}}(\mathbf{y} = (y_{e,1}, \dots, y_{e,K-1}), E = e \mid \mathbf{x})\\
&= \frac{1}{Z(\boldsymbol{\Theta}, \mathbf{x})}\exp(\sum_{k=1}^{K-1}(\boldsymbol{\theta}_{e,k}^T \mathbf{x} + b_{e,k})y_{e,k}),
\end{split}
\end{align}
with the normalizing constant \(Z(\boldsymbol{\Theta}, \mathbf{x}) = \sum_{i=1}^{K}\sum_{e'=1}^{E}\exp(\sum_{k=i}^{K-1}\boldsymbol{\theta}_{e',k}^T\mathbf{x} + b_{e',k})\) (we can sum over events since they are mutually exclusive). The predicted CIF is obtained from the PMF through

\begin{align}
\begin{split}
\hat{\mathrm{CIF}}_{\boldsymbol{\Theta}}(t, e, \mathbf{x}) = \sum_{k=1}^{t}\hat P_{\boldsymbol{\Theta}}(t_k, e, \mathbf{x})
\end{split}
\end{align}

For censored patients, since any event could occur first we marginalize over both time and events analogously to (\ref{eq:mtlr_censoring}):
\begin{small}
\begin{align}
P_{\boldsymbol{\Theta}}(T > t_j \mid \mathbf{x}) \propto \sum_{i=j}^{K-1}\sum_{e=1}^E\exp\Big(\sum_{k=i}^{K-1}(
\boldsymbol{\theta}_{e,k}^T \mathbf{x} + b_{e,k})y_{e,k}\Big),
\end{align}
\end{small}
with \(t_j\) again denoting the time of censoring.

The log-likelihood \(L(\boldsymbol{\Theta}, D)_{CR}\) can be derived analogously to the single event case (eq. (\ref{eq:mtlr_likelihood})) and has similar form to the DeepHit objective \cite{lee_deephit_2018}. In practice, we optimize the regularized likelihood \(L_{\mathrm{reg}}(\boldsymbol{\Theta}, D) = L(\boldsymbol{\Theta}, D)_{CR} + \frac{C_1}{2}\sum_{k=1}^{K-1}\Vert\boldsymbol{\theta}_k\Vert_2^2\), where \(C_1\) is a hyperparameter controlling the regularization strength and also the smoothness of predictions over time \cite{jin_using_2015}.

\begin{figure*}[ht]
    \centering
    \includegraphics[width=\textwidth]{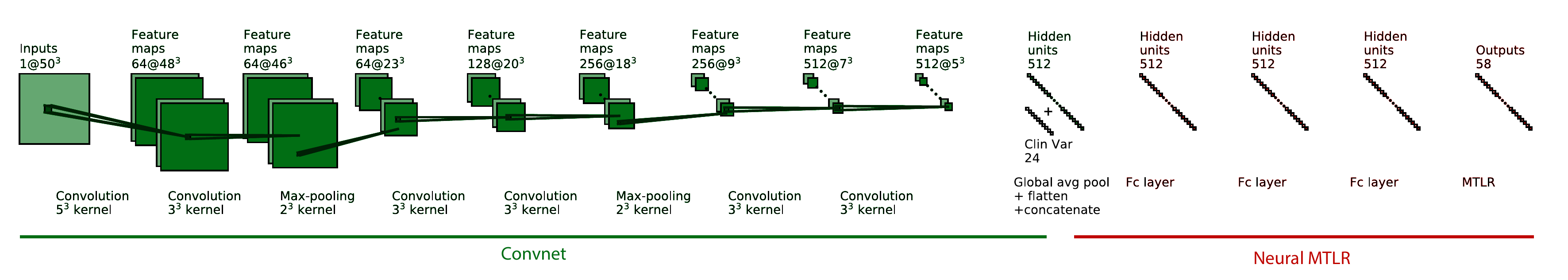}
    \caption{Deep-CR MTLR architecture.}
    \label{figure:diagram}
\end{figure*}
\subsection{Deep Cancer Prognostication With Multi-Modal Data}
We leverage the Neural MTLR framework \cite{fotso_deep_2018} to integrate the data across modalities. For EMR clinical data, categorical variables are one-hot encoded and continuous variables are normalized to be fed into the model. For medical images, we use a convolutional neural network to learn prognostic representations directly from voxels. The convnet output \(\mathbf{h}_{\mathrm{image}}\) is concatenated with the vector of clinical variables \(\mathbf{x}_{\mathrm{clinical}}\) and passed through shared fully-connected layers to learn shared features between modalities. Denoting the predicted survival probability at each timepoint and for each event as \(\hat P_{\boldsymbol{\Theta},\boldsymbol{\Phi}}(t, e, \mathbf{x}) \equiv P_{\boldsymbol{\Theta},\boldsymbol{\Phi}}(t_{k-1} < T \le t_k, E = e \mid \mathbf{x})\), we can obtain it using the MTLR head:
\[
\hat P_{\boldsymbol{\Theta},\boldsymbol{\Phi}}(t, e, \mathbf{x}) = \mathrm{MTLR}(f_{\boldsymbol{\Phi}}(\mathbf{x}_{\mathrm{clinical}}\Vert\mathbf{h}_{\mathrm{image}})),
\]
where \(\Vert\) denotes concatenation and \(f_{\boldsymbol{\Phi}}\) is a fully-connected neural network with parameters \(\boldsymbol{\Phi}\). The MTLR weights and the parameters of both convnet and the shared neural network are optimized jointly using gradient descent. 

\section{Experiments}

\subsection{RADCURE Dataset}
The RADCURE dataset consists of 2552 HNC patients treated with radiotherapy or radiochemotherapy at Princess Margaret Cancer Centre between 2005--2018. For each patient, pre-treatment computed tomography (CT) image with the tumour manually delineated by an experienced radiation oncologist was available alongside EMR data (age at diagnosis, sex, disease site and stage, overall health status human papillomavirus (HPV) infection status, radiation dose and use of chemotherapy). More in-depth information of the dataset can be found at \cite{kazmierski_machine_2021}. We split the dataset into 60\% training, 10\% validation and 30\% test sets by date of diagnosis (i.e. all the test set patients were diagnosed after the most recently-diagnosed training patient) to simulate a prospective study. The patients were followed up for at least 2 years (or until death if occurred before). There are 2 competing events: death from primary cancer or other causes (e.g. due to comorbidities or another cancer). Because of the temporal splitting scheme used, there are fewer events in the test set than in the training set due to shorter follow-up time. Table \ref{tab:data_overview} shows an overview of the training and test sets.
\begin{footnotesize}
\newcommand{\rowgroup}[1]{\hspace{-1em}#1}
\begin{table}[h!]
\centering
\begin{tabular}{>{\quad}lll}
\toprule
& \textbf{Training/validation} & \textbf{Test} \\
\midrule
\rowgroup{\# of patients} & 1802 & 750\\

\rowgroup{Outcome}\\
Alive / Censored & 1065 (59\%) & 609 (81\%)\\
Death (Cancer)   & 342  (19\%) & 102 (14\%)\\
Death (Other)    & 395 (22\%) & 39  (5\%)\\




\bottomrule
\end{tabular}
\caption{Event distribution in training/validation and test sets.}
\label{tab:data_overview}
\end{table}
\end{footnotesize}
\subsection{Deep-CR MTLR Architecture}
Our Deep-CR MTLR architecture is shown in \ref{figure:diagram}, and was chosen after testing various architectures and kernel sizes on the same  training/validation data. The final architecture's main benefit is its simplicity, which increases generalizability and decreases the risk of overfitting. The network takes \(50^3\) voxel volumes centered around the primary tumor as input, followed by 2 3D convolutional blocks, each consisting of 2 pairs of convolution and ReLU activation, and \(2^3\) max pooling with stride 2. Finally, the 512 channels flattened by global average pool on each channel, resulting in a flattened vector of size 512. Clinical variables are concatenated to the flattened vector, resulting in a vector of size 536. This vector is passed to the shared neural MTLR, with 3 hidden layers and ReLU activations which returns the predicted joint probability of time and event.

To investigate the added prognostic value of multi-modal representations, we compare the performance of the combined model with the shared network applied individually to clinical data or convnet outputs alone, and a linear MTLR using clinical variables only.

\subsection{Performance Metrics}
To evaluate and compare the prognostic performance of the models, we used two performance metrics. The \textit{cause-specific concordance index} (\(C\)-index) \cite{harrell_multivariable_1996} is a ranking metric evaluating the ability to assign higher predicted risk to patients experiencing the event at and earlier time. We also used \textit{cause-specific area under the receiver operating characteristic curve} (AUROC) evaluated at 2 years, which is a threshold-free metric analogous to the \(C\)-index for binary classification. The 2-year AUROC lets us evaluate the model performance in identifying particularly high-risk patients, while the \(C\)-index provides an indication of lifetime risk prediction accuracy. We compute the AUROC between the predicted CIF evaluated at 2 years and the indicator variable \(\mathbf{1}\{T^{(j)} \le 2 \land E^{(j)} = e\}, j = 1, \dots, N\) for each event \(e\). The \(C\)-index is computed using the predicted lifetime risk score, obtained by aggregating the predictions over all timepoints.

\subsection{Training Details}
The model was implemented using the PyTorch package \cite{paszke_pytorch_2019} on the PyTorch Lightning framework \cite{falcon2019pytorch}. We trained the network using the Adam optimizer \cite{kingma_adam:_2014} with learning rate \(10^{-4}\). The hyperparameters, including the number of hidden layers in the shared network and \(C_1\) regularization constant were selected according to performance on the pre-defined validation set. We used separate regularization for the neural network parameters \(\boldsymbol{\Phi}\) and MTLR parameters \(\boldsymbol{\Theta}\) as we found that the latter usually need stronger regularization to maintain smoothness over time intervals. The number of time discretization intervals was selected as \(\sqrt{\text{\# of training patients}}\).

\subsection{Model Performance}
Table \ref{tab:results} shows the 2-year AUROC and \(C\)-index of Deep-CR MTLR alongside modality-specific models. The best performance was achieved by the multi-modal model with 3  hidden layers (AUROC of 0.774 and 0.781, \(C\)-index of 0.788 and 0.761), for cancer and other causes, respectively. Compared to each of the modalities alone, combining imaging and clinical data leads to substantial improvement in both cancer-specific and other cause survival prediction, particularly in terms of the lifetime risk \(C\)-index. This aligns with our hypothesis that additional information present in medical images can be better exploited in predicting cancer-specific survival. Additionally, all of the deep models performed better than the linear MTLR, highlighting the role of learning non-linear interactions in accurate prognosis. 

\begin{footnotesize}
\begin{table}[h!]
\centering
\begin{tabular}{@{}lllll@{}}
\toprule
 & \multicolumn{2}{c}{\textbf{2-year AUROC}} & \multicolumn{2}{c}{\textbf{\(C\)-index}}\\
 \cmidrule(lr){2-3}
 \cmidrule(lr){4-5}
\textbf{Model} & Cancer & Other & Cancer & Other \\
\midrule
Linear MTLR (clinical)    & 0.740 & 0.724 & 0.747 & 0.679  \\
Convnet (images)        & 0.742 & 0.740 & 0.746 & 0.707 \\
N-MTLR (clinical)           & 0.768 & 0.755 & 0.756 & 0.726 \\
\textbf{\textit{Deep-CR MTLR} (both)} & \textbf{0.774} & \textbf{0.781} & \textbf{0.788} & \textbf{0.761}  \\
\bottomrule
\end{tabular}
\caption{Performance metrics of combined multi-modal vs benchmark models on test set}
\label{tab:results}
\end{table}
\end{footnotesize}

\section{Conclusions}
In this study, we have presented a novel method for integrating multi-modal data to predict individual survival in cancer in the presence of competing risks. Deep-CR MTLR performs better than the individual modalities alone, particularly in cancer-specific survival prediction, indicating that the new architecture is able to leverage both modalities together to predict future outcomes. Our model is a promising new tool for oncologists, enabling them to exploit routinely collected clinical and imaging data to make more informed decisions tailored to the individual patient.\\

In this paper, we do not address the performance of all-cause survival predictions of the data, as this is already comprehensively addressed in \cite{kazmierski_machine_2021}. It is interesting to understand Deep-CR MTLR's advantages and disadvantages compared to Cox-PH based methods such as DeepSurv \cite{katzman_deepsurv_2018} and deep learning based competing risks DeepHit \cite{lee_deephit_2018}, which we leave for future work.

\section{Acknowledgements}
The authors would like to thank Dr. Scott Bratman and Dr. Andrew Hope for assistance with data collection and helpful discussions. MK is supported by the Strategic Training in Transdisciplinary Radiation Science for the 21st Century Program (STARS21) scholarship.

\section{Code Availability}
The source code of the study is available at: \url{https://github.com/bhklab/aaai21_survival_prediction}

\bibliographystyle{aaai}
\bibliography{ms.bib}

\end{document}